# Beyond feature attribution: Recovering latent interaction structures in machine learning models via localized CBFI

Sejong Oh[1]

[1] Department of Software Science, Dankook University, Gyeonggi-do, Republic of Korea

Corresponding Author:
Sejong Oh[1]
152, Jukjeon-ro, Yongin, 16890, Republic of Korea
Email address: sejongoh@dankook.ac.kr

## Abstract

**Background:** Understanding individual predictions in complex tabular models remains challenging, as existing additive frameworks like SHAP and LIME often obscure structural logic by absorbing nonlinear interactions into individual feature scores. Exposing these latent interaction structures is essential for automated auditing in high-stakes environments where understanding the logic of model behavior is as essential as predictive accuracy.

**Methods:** To address this limitation, we introduce the localized case-based feature importance (CBFI) framework. Unlike traditional numerical attribution, our method decomposes instance-specific predictions into structurally distinct components, effectively separating independent main effects (*G1*) from pure interactions (*G4*) to capture explicit feature synergies and provide a principled approach for diagnosing feature combinations.

**Results:** Comprehensive experiments on wine quality classification and medical insurance regression demonstrate that localized CBFI successfully disentangles complex interactions hidden by additive models. Furthermore, fidelity tests and synthetic ground-truth benchmarks confirm that the identified interactions correspond precisely to the nonlinear structural dependency patterns embedded in complex architectures, including gradient boosting and deep neural networks, thereby successfully shifting the explainable AI paradigm from simple numerical attribution to the recovery of interaction-driven contributions.

## Introduction

The paradigm of modern artificial intelligence research has evolved beyond maximizing predictive performance (accuracy) to emphasize interpretability, where the decision-making process of models can be justified in terms understandable to humans (Adadi & Berrada, 2018; Miller, 2019). This shift is particularly important in high-stakes domains such as financial actuarial analysis, medical diagnosis, and public policy decision-making, where tabular data are commonly used (Rudin, 2019**)**. In these settings, complex models, such as artificial neural networks (ANNs) and gradient boosting decision trees (GBDTs), have demonstrated superior predictive performance (Arrieta et al., 2020). However, their inherent black-box characteristics have limited their adoption in real-world practices. Consequently, local explanation techniques, which quantify the influence of each feature on the prediction of a specific input instance, have emerged as key approaches for improving the transparency and trustworthiness of machine learning models (Guidotti et al., 2018; Arrieta et al., 2020).

## Motivation and problem statement

Representative approaches for local model explanations include Shapley additive explanations (SHAP) (Lundberg & Lee, 2017) and local interpretable model-agnostic explanations (LIME) (Ribeiro, Singh & Guestrin, 2016). These methods aim to interpret the prediction of a complex machine learning model at a specific input instance by expressing the prediction in terms of the contributions from individual features. In these approaches, the model output for a given instance is decomposed into a set of feature-level attribution values whose additive combination approximates or reproduces the original prediction.

However, this additive assumption may fail to adequately capture the complex interactions among variables that frequently arise in real-world data (Molnar, Casalicchio & Barch, 2020; Slack et al., 2020). In the practical decision processes of machine learning models, certain features may attenuate the influence of others, producing suppression effects, whereas some features may exert a significant influence only when combined with other features, forming synergistic interaction patterns. Additive explanation models tend to implicitly absorb these interaction components into the attribution values assigned to individual features. Consequently, users may interpret the model as evaluating each feature independently, potentially overlooking the nonlinear decision mechanisms embedded in the model.

To address this limitation, this study investigates the localization of case-based feature importance (CBFI) (Oh, 2022), a method originally designed to analyze the stable contribution structures of models at the global level. By projecting the structural contribution decomposition identified in global analysis onto individual prediction instances, the proposed approach aims to explicitly distinguish between independent feature influence and interaction-driven effects. Clarifying this distinction constitutes the central motivation for this study.

## Related work and contributions

Understanding how machine learning models generate predictions has become a central challenge in explainable AI (XAI) (Gilpin et al., 2018). Existing local explanation methods primarily aim to quantify the contributions of individual features to specific predictions. These approaches can be broadly categorized into surrogate-model-based explanations and attribution-based explanations.

Surrogate model approaches attempt to approximate the behavior of complex models using simpler interpretable models in the neighborhood of a target instance. A representative method is LIME, which generates perturbed samples around a given instance and fits a locally weighted linear model to approximate the prediction boundary. While such surrogate models provide intuitive explanations, their fidelity strongly depends on the quality of the local sampling and the stability of the surrogate approximation.

In contrast, attribution-based approaches directly estimate feature contributions without constructing surrogate models. The most widely used method is SHAP, which derives feature importance values using the Shapley value formulation from cooperative game theory. SHAP computes the expected marginal contribution of each feature across all possible coalitions, thus providing theoretically grounded additive explanations that satisfy properties such as efficiency and consistency.

Despite their popularity, additive attribution frameworks have a fundamental limitation in that they assume that model predictions can be decomposed into the sum of individual feature contributions. However, modern machine learning models frequently rely on nonlinear feature interactions, where predictive signals emerge only through the joint influence of multiple variables (Tsang et al., 2018). In such cases, additive decomposition may absorb the interaction effects into individual feature scores, thereby obscuring the interaction-driven decision mechanisms used by the model.

To address this limitation, several approaches have attempted to quantify feature interactions more explicitly. For example, SHAP interaction values (Lundberg et al., 2020) extend the SHAP framework by decomposing contributions into pairwise interaction terms, whereas Friedman's H-statistic (Friedman & Popescu, 2008) measures the overall Fstrength of interactions in predictive models. Although these techniques enable quantitative analysis of interactions, they face two important challenges. First, the number of possible interactions increases combinatorially with the number of features, leading to significant computational overhead. Second, interaction scores are typically interpreted alongside additive feature importance values, which can make it difficult to clearly isolate the structural role of interactions in a model's decision process.

More fundamentally, existing interaction metrics primarily quantify interaction magnitude rather than disentangling the interaction-driven structures that govern the model's behavior. Consequently, they provide limited insight into whether a prediction depends on a synergistic combination of features or independent feature effects.

This limitation motivates a different perspective on explanation. Instead of distributing importance across features, an explanation method should aim to identify the structural dependency patterns utilized by the model during prediction.

In this study, we built on the CBFI framework, which decomposes prediction outcomes into mutually exclusive structural components representing distinct roles in the model's decision structure. By extending this framework to the instance level, we introduced localized CBFI, a structural decomposition approach designed to expose the structural dependency patterns embedded in machine learning models.

Rather than interpreting interactions as residual effects within additive explanations, the proposed method explicitly separates independent feature contributions from pure synergistic interactions, thereby enabling the identification of the latent interaction structure underlying the model predictions. This perspective shifts the focus of XAI from importance attribution to exposing structural dependency patterns, providing a more faithful representation of how complex models combine features to generate predictions.

Therefore, understanding model behavior requires moving beyond feature attribution toward **exposing the structural dependency patterns** embedded in machine learning models.

# Materials & Methods

## Review of global CBFI decomposition

The CBFI framework provides a rigorous methodology to decompose the overall effect of a feature into distinct structural components. Unlike conventional importance measures that provide a single unitary score, CBFI explicitly partitions feature influences into independent and synergistic components. In this study, the independent component (*G1*) is referred to as the "Main Effect," extending the concept originally termed "Feature Power" in Oh (2022).

For a classification model, CBFI observes the correctness of predictions across different datasets: the original dataset (*DS*), the dataset where the complementary features of $F_x$ are permuted ($DS(F_x)$), and the dataset where $F_x$ itself is permuted ($DS(F_{x(-)})$). Based on these observations, the correctly predicted cases in the original model were classified into four mutually exclusive and exhaustive groups.

- Group 1 (*G1*, Main Effect): Instances where only the target feature $F_x$ provides sufficient information for a correct prediction, whereas the remaining features $F_{x(-)}$ do not. This represents the independent "Main Effect" of the feature.
- Group 2 (*G2*, Alternative Contribution): Instances where $F_x$ fails to predict correctly but the remaining features $F_{x(-)}$ succeed. This represents the predictive power inherent in the complementary feature set, independent of $F_x$.
- Group 3 (*G3*, Common Contribution): Instances where both $F_x$ and $F_{x(-)}$ independently provide sufficient information for a correct prediction.
- Group 4 (*G4*, Interaction): Instances where neither $F_x$ nor $F_{x(-)}$ alone can produce a correct prediction; however, their combination enables the model to produce a correct outcome in the full model.

In this framework, the total feature importance (*Imp*) of $F_x$ is mathematically defined as the sum of its main effect (*G1*) and cooperative interaction (*G4*).

$$Imp(F_x) = FP(F_x) + Int(F_x, F_{x(-)}) = (n(G1) + n\,(G4))/NI \quad (1)$$

where *FP*() denotes the main effect of a feature, *Int*() denotes the interaction between a feature and other features, *n*() denotes the number of cases in each group, and *NI* denotes the total number of instances.

For regression models, the grouping logic is extended by analyzing the change in the size of the prediction errors (*diff*). A feature is considered to have a positive contribution if its presence reduces the prediction error compared with when it is permuted. Specifically, *G4* (interaction) in the regression is identified when both $F_x$ and $F_{x(-)}$ individually result in higher errors than the full model, indicating that the joint presence of these features is necessary for an accurate estimation. This decomposition allows the CBFI to capture cooperative dependencies between features that cannot be explained by their independent effects.

## Instance-level contribution decomposition

The overall framework for decomposing model predictions into independent main effects (*G1*) and synergistic interactions (*G4*) at the instance level is illustrated in Figure 1.

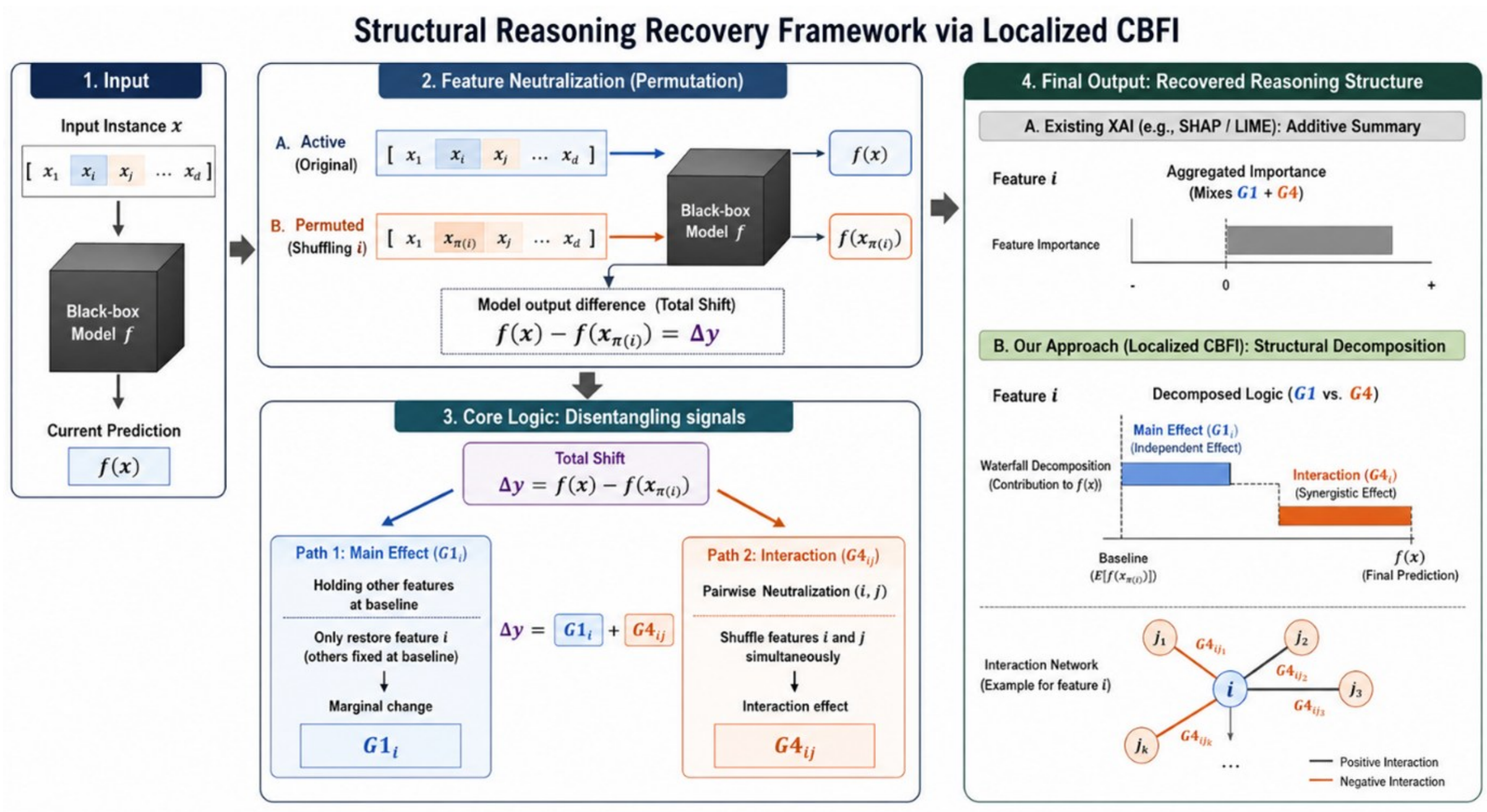

**Fig. 1**. Conceptual framework of localized CBFI for exposing structural dependency patterns. The process illustrates (1) input of an instance into the black-box model, (2) feature neutralization via permutation shuffling, (3) decomposition of the predictive shift (Δy) into independent (*G1*) and interaction-driven (*G4*) components, and (4) the final output providing structural diagnostics compared to traditional additive summaries.

Localized CBFI is established as a structural decomposition framework that extends the contribution-based decomposition logic of global CBFI to individual predictions. Rather than merely providing local attribution, this framework was designed to decompose model predictions into independent contributions and interactions, thereby exposing the structural dependency patterns of feature contributions that are typically obscured by additive approximations.

Localized CBFI decomposes the instance-level feature influence into two complementary components: the main effect (*G1*), which represents the independent contribution of a feature, and interaction (*G4*), which represents the contribution that emerges only from the joint presence of multiple features. Through this decomposition, the framework reveals the structural dependencies governing the decision structure of the model.

To isolate the structural contribution of a specific feature, the framework employs a permutation-based sampling strategy rather than relying on a single static reference value. By randomly shuffling the feature values across the dataset, the proposed method disrupts the logical association between the target feature and the specific prediction instance, while preserving the feature's marginal distribution. This permutation process effectively neutralizes specific feature information, allowing the model to generate a set of stochastic counterfactual predictions. These outputs represent the model's behavior when the target feature is "active" in terms of its distribution but "decoupled" from its original instance-level logic. This nuanced neutralization provides the necessary baseline for localized CBFI to rigorously decompose the total predictive signal into independent main effects (*G1*) and synergistic interactions (*G4*), ensuring that the identified latent interaction structure is not biased by the out-of-distribution artifacts that are often introduced by fixed-point replacements.

The main effect ($G1_i$) at the individual instance level represents the independent influence of the *i*-th feature on the prediction outcome, isolated from any interactions with other variables. Unlike conventional methods that rely on static reference points, we define *G1* as the difference between the original prediction and the expected prediction under permutation:

$$G1_i(x) = f(x) - \mathbb{E}_{\pi \in \Pi}[f(x_{\pi(i)}] \tag{2}$$

where $\mathbb{E}_{\pi \in \Pi}[\mathrm{f}(x_{\pi(i)}]$ denotes the model's expected prediction for instance x when the *i*-th feature is neutralized via permutation shuffling. This ensures that the baseline reflects the marginal distribution of the feature rather than an arbitrary fixed value.

The interaction ($G4_{ij}$) is defined as the component representing interaction-driven decision mechanisms that emerges from the joint presence of the *i*-th and *j*-th features. Unlike traditional interaction metrics that merely quantify nonlinear residuals, $G4_{ij}$ is specifically designed to identify the structural dependency patterns inherent in the model's internal mechanisms. Formally, the interaction between two features *i* and *j* is calculated as the change in prediction when both features are simultaneously neutralized via permutation after subtracting their respective independent main effects ($G1_i$ and $G1_j$):

$$G4_{ij}(x) = f(x) - \mathbb{E}_{\pi \in \Pi}[f(x_{\pi(i,j)}] - (G1_i(x) + G1_j(x)) \tag{3}$$

This pairwise formulation allows the framework to map the model's latent interaction network. By aggregating these pairwise synergistic effects, we can derive the total interaction-driven importance of a specific feature i, thereby providing a comprehensive view of how the feature participates in the complex interaction-driven decision mechanisms of the model.

For regression models, decomposition was performed based on changes in prediction errors (*diff*), allowing the framework to identify moderating mechanisms and essential interdependencies between variables. By explicitly separating independent contributions from interactions, the proposed approach addresses a key limitation of additive explanation models, which often absorb nonlinear dependencies into individual feature importance scores, thereby masking the underlying interactions within the model's decision structure.

Finally, the proposed framework is model-agnostic, allowing it to be applied to a wide range of predictive models for tabular data while preserving the structural interpretation of feature interactions.

## Interpretative framework of localized metrics

To ensure consistent interpretation of the experimental results, we established a technical framework for evaluating the sign and magnitude of the localized metrics.

### Directionality of local main effect (*G1*)

The local main effect, $G1_i(x)$, quantifies the independent drive of feature *i*. A positive value indicates that the feature independently pushes the model prediction higher relative to the reference state, whereas a negative value signifies an independent downward influence. Unlike additive attributions in SHAP or LIME, *G1* is strictly isolated from interaction components, representing the "pure" influence of a single variable.

### Categorization of local interaction (*G4*)

The local interaction, $G4_{ij}(x)$, represents the structural dependency between features that emerge beyond their independent contributions. Rather than merely indicating the magnitude of the interaction, *G4* reveals how multiple variables jointly participate in the model's decision structure. Therefore, the sign of *G4* provides insight into the structural role that feature combinations play in shaping predictions.

- Synergistic Interaction ($G4 > 0$)

  A positive value indicates that the predictive signal emerges from the cooperative presence of features i and j. In this case, the joint contribution of the features exceeds the sum of their independent main effects, suggesting that the model relied on synergistic interactions to generate the prediction.

- Regulatory or Suppressive Interaction ($G4 < 0$)

  A negative value indicates that the joint presence of the features constrains or moderates their independent influences. Such patterns often correspond to regulatory interactions in which one feature modulates or suppresses the effects of another. In regression tasks, these negative interactions can reveal the mechanisms that stabilize or limit the predicted output of a model.

### Scale of absolute contribution

A distinctive characteristic of the localized CBFI framework is its reliance on non-normalized absolute scales (e.g., actual currency units in regression or direct probability shifts in classification). This allows for a direct "meaning-based" interpretation of importance, providing a more intuitive grasp of how much a specific feature set alters the model's decision in real-world units.

## Case-based sampling strategy

A major technical challenge in analyzing feature interactions is the "combinatorial explosion" problem, where the number of possible combinations grows exponentially as the number of features ($n$) increases. Existing game-theory-based methods incur exponential computational costs because they must evaluate all possible feature subsets (coalitions). To address this bottleneck, the localized CBFI framework incorporates case-based analysis—the core philosophy of the original CBFI—into its sampling process.

Rather than an exhaustive search, the sampling strategy focuses on tracking "significant prediction changes" that characterize the decision structure for a specific instance. For feature neutralization, the framework adopts a permutation-based approach using random samples from the background dataset. Instead of relying on fixed statistics, such as the median, this method measures contributions by substituting target features with random values drawn from the data distribution, thereby preserving the structural integrity of the manifold learned by the model.

The case-based sampling strategy proceeds through the following steps:

- **Region Definition and Candidate Selection**: A local region is established within the feature

space, and the primary feature candidates that exert a dominant influence on the instance prediction are identified.

- **Group Classification and Contribution Recording**: Based on the identified features, the prediction cases were classified into mutually exclusive groups (*G1* to *G4*), and the magnitude of prediction change at each stage was recorded.
- **Ensuring Statistical Stability**: To minimize the variance from random sampling, the final importance metrics were calculated by averaging the contribution scores across multiple iterations. This robust averaging process reduces the noise and ensures the reliability of the explanation.

This strategy maintains the computational complexity at a manageable level while precisely reconstructing the nonlinear interactions within the model. Consequently, localized CBFI offers a practical advantage by identifying synergistic and regulatory mechanisms between features more efficiently than conventional additive metrics, even in high-dimensional tabular data environments.

## Experimental Setup

This section describes the experimental environment used to evaluate the validity and effectiveness of the proposed localized CBFI framework. We conducted experiments across two distinct predictive tasks, classification and regression, to analyze how the framework identifies feature interactions in diverse data environments.

### Datasets

We utilized two public datasets characterized by distinct domain attributes and complex nonlinear relationships. Table 1 summarizes the statistical properties of each dataset. These datasets were used in their raw format without any preprocessing.

- Wine Quality Dataset (https://archive.ics.uci.edu/dataset/186/wine+quality) (Cortez et al., 2009): This dataset consists of red wine samples with 11 chemical properties, including acidity, residual sugar, and alcohol. It is particularly suitable for analyzing regulatory mechanisms (*G4*) because the final quality grade is often determined by the delicate balance and interdependence between multiple components rather than the independent influence of a single feature.
- Medical Insurance Dataset (https://github.com/stedy/Machine-Learning-with-R-datasets/blob/master/insurance.csv) (Lantz, 2013): This dataset predicts annual medical charges based on individual attributes, including age, BMI, and smoking status. It possesses a strong nonlinear dependency structure, where the combination of "Smoker" status and numerical variables (e.g., age or BMI) creates an exponential synergy effect on the prediction. This makes it an ideal candidate for verifying whether localized CBFI can decouple essential interdependencies from the main effect.

### Model selection and training

To demonstrate the model-agnostic nature of the localized CBFI framework, we adopt the "Selection and Focus" strategy involving two architecturally different models:

- **Random Forest (RF)** (Breiman, 2001)**:** As the primary model for detailed quantitative analysis,

the random forest is employed because of its superior performance on tabular data and inherent ability to capture complex feature interactions through hierarchical splits.

- **Multi-Architecture Validation (Diverse Baselines):** To rigorously validate the model-agnostic nature and generalizability of the localized CBFI framework, we conducted extensive testing across five different model architectures, including artificial neural networks (ANN) (Rumelhart, Hinton & Williams, 1986; Grinsztajn, Oyallon & Varoquaux, 2022), gradient boosting machines (XGBoost (Chen & Guestrin, 2016) and CatBoost (Prokhorenkova et al., 2018)), TabNet (Arık & Pfister, 2021), and support vector machines (SVM (Cortes & Vapnik, 1995)). Our experimental results confirmed that the interaction structures (*G4*) identified by the framework remained consistent across these diverse decision mechanisms. To ensure a focused and concise narrative, the following sections present detailed analyses centered on the RF results, whereas comprehensive benchmarking data for all other architectures are provided in Appendix A.3.

The numerical features were standardized using StandardScaler, and each dataset was split into training and test sets in an 8:2 ratio.

### Baseline methods and XAI configuration

The performance of localized CBFI was evaluated against SHAP and LIME, which are the current state-of-the-art methods for local explanation.

- SHAP: KernelSHAP and TreeSHAP were used to calculate Shapley values, which represent the average marginal contribution across feature coalitions.
- LIME: Local feature importance was derived by training locally weighted linear surrogate models on 5,000 perturbed samples.
- Localized CBFI: To compute local *G1* (main effect) and *G4* (interaction), we implemented the case-based sampling strategy defined in Section 2.4. Following the permutation principle of the original CBFI framework, feature neutralization was performed by replacing target features with values randomly sampled from the background dataset. To ensure statistical stability and mitigate the effect of random variance, the final contribution scores were calculated by averaging the results over multiple sampling iterations ($n = 100$).

### Structural XAI Evaluation Metrics

It is crucial to note that a direct quantitative comparison of importance scores between localized CBFI and traditional additive frameworks (e.g., SHAP, LIME) is fundamentally inappropriate. Additive methods aim to distribute predictive influence as marginal attributions, whereas localized CBFI explicitly decomposes feature combinations to recover latent interaction structures. Because they operate on completely different explanatory paradigms, comparing their raw attribution scores as a measure of "superiority" is structurally flawed.

Therefore, rather than relying on direct numerical attribution comparisons, we utilize four specialized evaluation metrics designed to assess the **structural truthfulness**, **discriminative resolution**, and **robustness** of the explanations:

1. **Explanatory Fidelity (Structural Degradation)**: Fidelity evaluates how faithfully an explanation identifies the actual decision-making criteria of a black-box model. We measure this by iteratively neutralizing (permuting) the top-$k$ features identified as important and tracking the degradation of the model's performance.

- For classification, we plot the decline in the predicted probability of the correct class:

$$Fidelity_{clf}(k) = f(x)_{|correct-class|} - f(x_{\pi(1:k)})_{|correct-class|} \tag{4}$$

- For regression, we plot the shift in absolute error relative to the ground truth:

$$Fidelity_{reg}(k) = |f\left(x_{\pi(1:k)}\right) - y| - |f(x) - y| \tag{5}$$

A steeper drop (or larger shift) indicates that the explanation model has successfully captured the "binding requirements" essential to the model's decision structure, rather than just independent marginal effects.

2. **Interaction Sensitivity (Signal-to-Noise Ratio)**: To evaluate how cleanly an interaction metric isolates true synergistic relationships from additive noise without unfair quantitative comparisons to SHAP, we define **Sensitivity** using a synthetic ground-truth environment. It is the ratio of interaction strength assigned to truly interactive features (the "Signal") to the strength assigned to purely additive features (the "Noise"):

$$Sensitivity = \frac{\mathbb{E}[|G_4(x_{interactive})|]}{\mathbb{E}[|G_4(x_{additive})|]} \tag{6}$$

A higher Sensitivity ratio demonstrates superior discriminative power, proving that the metric successfully isolates synergies without leaking additive effects.

3. **Explanatory Divergence and Alignment**: Instead of using the **Pearson Correlation Coefficient (*R*)** to claim quantitative superiority, we utilize it as an analytical tool to demonstrate *where* additive models fail.

$$R = \frac{\sum(I_{CBFI} - \bar{I}_{CBFI})(I_{baseline} - \bar{I}_{baseline})}{\sqrt{\sum(I_{CBFI} - \bar{I}_{CBFI})^2 \sum(I_{Baseline} - \bar{I}_{BaselineCBFI})^2}} \tag{7}$$

While high correlation (alignment) acts as a sanity check in simple linear regimes, a low correlation (divergence) in complex regression tasks empirically proves the structural distortion of additive frameworks, highlighting regions where localized CBFI restores the masked interaction logic.

4. **Sampling Stability (Coefficient of Variation, CV)**: Because the case-based sampling strategy relies on a stochastic permutation scheme, we must ensure that the structural explanations are numerically stable. We quantify this using the Coefficient of Variation (CV):

$$CV = \frac{\sigma(G_4)}{\mu(G_4)} \tag{8}$$

where $\sigma(G_4)$ and $\mu(G_4)$ represent the standard deviation and mean of the localized interaction strength ($G_4$) across 50 independent runs. A CV below the 10% threshold confirms that the identified structural dependencies are robust against sampling noise.

## Implementation Details

To ensure the reproducibility of the experiments, we provided the following details regarding the hardware and software environments used in this study.

- **Hardware Specifications**: All experiments were conducted on a system equipped with an Intel Core Ultra processor 5 (with NPU) and 16 GB of RAM. The system utilized Intel Iris Xe Graphics to assist with general-purpose computing tasks using the Windows 11 operating system.
- **Software Environment**:
  - **Programming Language**: The entire framework and experimental pipelines were implemented using Python (v3.14+).
  - **Machine Learning Libraries**: Scikit-learn (v1.8.0) was used to train and evaluate the random forest model.
  - **XAI Frameworks**: The baseline explanations for SHAP and LIME were generated using the official shap (v0.51.0) and lime (v0.2.0.1) Python packages, respectively.

**Data Manipulation and Visualization**: Data preprocessing and matrix operations were performed using pandas and numpy (v2.4.3), whereas matplotlib (v3.10.8) was employed to generate analytical visualizations.

## Results

This section presents the experimental results of the localized CBFI framework for wine quality classification and medical insurance charge regression. We evaluated the performance of the framework based on quantitative validity, explicit decomposition of feature effects, identification of regulatory mechanisms, and explanatory fidelity.

### Quantitative validity and explanatory consistency

To verify the mathematical validity of localized CBFI, we performed a correlation analysis using established local explanation models. The results of the correlation analyses are summarized in Table 2.

- **Convergent Validity:** In the wine quality dataset, the CBFI importance scores were highly correlated with SHAP (0.726) and LIME (0.757). In the insurance charge regression, the correlation with absolute SHAP values reached **0.9311**, demonstrating that localized CBFI aligns well with the attribution logic of state-of-the-art models.
- **Stability of Explanation:** Notably, in the insurance dataset, while LIME and SHAP showed a negative correlation (-0.6029), indicating conflicting explanations, CBFI maintained a stable positive correlation with LIME (0.7561). This suggests that localized CBFI captures the underlying structural truth of the model more robustly when additive approximations diverge.

### Explicit decomposition of main effect (*G1*) and interaction (*G4*)

The most significant academic contribution of localized CBFI is its ability to mathematically decompose individual feature importance into independent main effect (*G1*) and interaction (*G4*) resulting from feature combinations. This granular approach reveals the underlying dependency structures within the model that additive models such as SHAP or LIME fail to capture by presenting only a single integrated value.

#### Wine quality dataset: Identifying interaction and regulatory mechanisms

The wine quality classification model clearly distinguished the different roles of chemical properties in the model's decision-making process. Figure 2 summarizes the result.

- **Pure Interaction Features (alcohol, total sulfur dioxide):** For the alcohol variable, the total importance (0.42) was entirely composed of the interaction (*G4*) component, with no observed main effect (*G1*). This suggests that alcohol content does not determine quality in isolation but exerts a significant influence only in "balance" with other chemical components. Similarly, the contribution of total sulfur dioxide was formed exclusively by interaction (*G4*: 0.21), validating its role as a context-dependent feature.
- **Mixed Contribution Features (sulphates, volatile acidity):** In contrast, sulphates exhibited both a main effect (*G1*: 0.23) and an interaction (*G4*: 0.06) simultaneously. This result indicates that the sulfate content provides an independent contribution to the quality grade, while also generating additional synergy through chemical combinations with other components.
- **Background Information Features (residual sugar, citric acid):** While features such as residual sugar and citric acid showed importance scores (*G1*+*G4*) near zero, their *G2* (Alternative Contribution) and *G3* (Common) metrics were exceptionally high, exceeding 0.90. This demonstrates that although these variables are not decisive for the final classification of a specific instance, they are deeply involved in the model's internal logic as foundational background data for stable predictions.

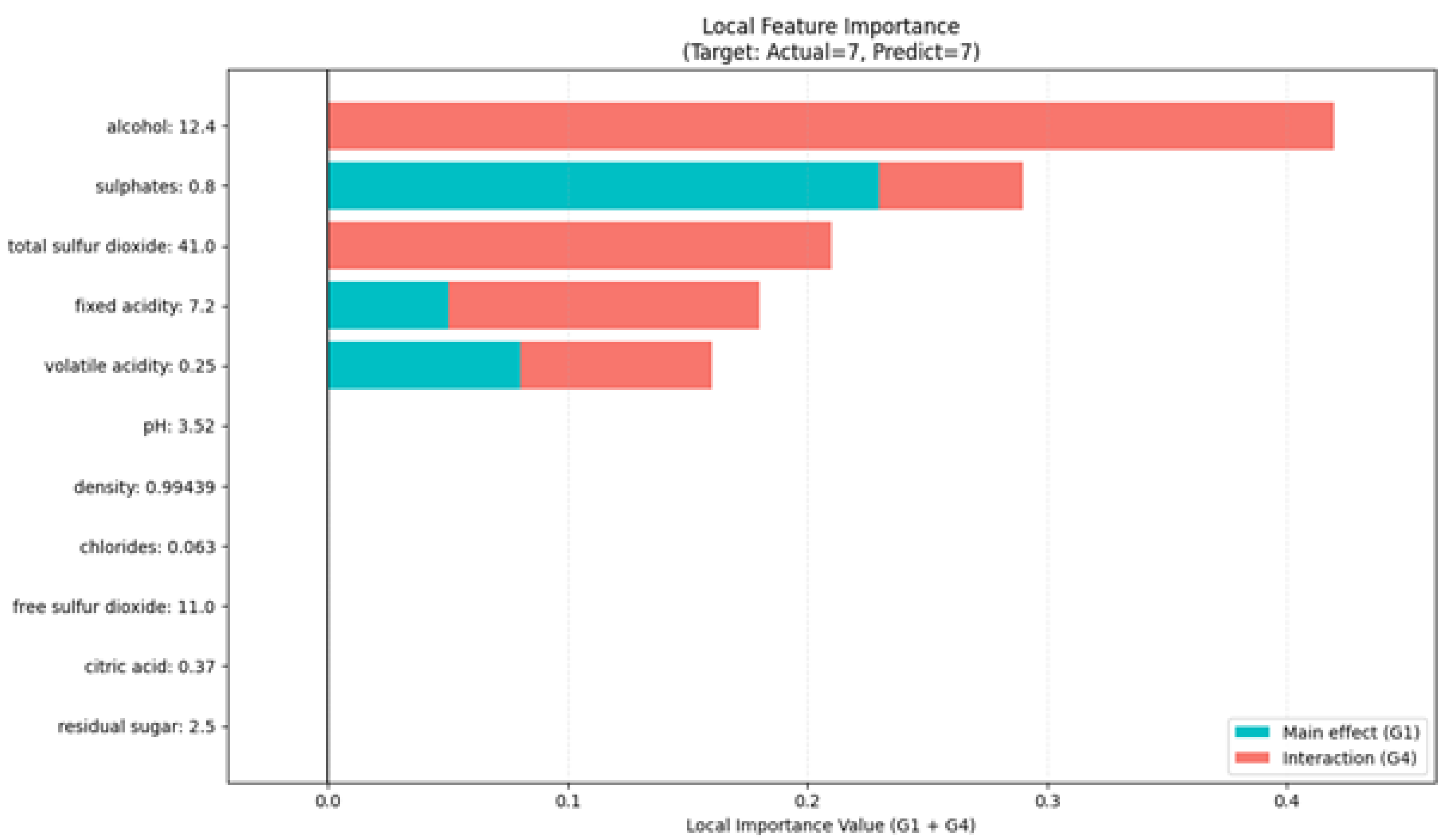

**Fig. 2.** Local feature importance for wine quality dataset (*G1* vs. *G4* decomposition).

### Medical insurance dataset: Restoring explosive synergies

In the insurance charge regression model, powerful nonlinear dependencies between specific feature pairs are evident through numerical decomposition (Figure 3).

- **Dominance of Interaction Components:** For the highest-ranked features, smoker (6017.25) and

age (5594.36), the vast majority of their contributions were derived from interaction (*G4*). Notably, the smoker variable showed almost no main effect (*G1*), implying that the model does not penalize "smoking" itself in a vacuum. Instead, localized CBFI accurately pinpointed the nonlinear synergy, in which charges escalate exponentially when a smoker is older or has a high BMI.

- **Contrast with Additive Models:** While SHAP merely assigns high importance scores to these variables, localized CBFI distinguishes whether these scores stem from independent influences or amplification through coupling. This provides actuaries and medical experts with a much richer and more reliable evidentiary basis for high-stakes decision making.

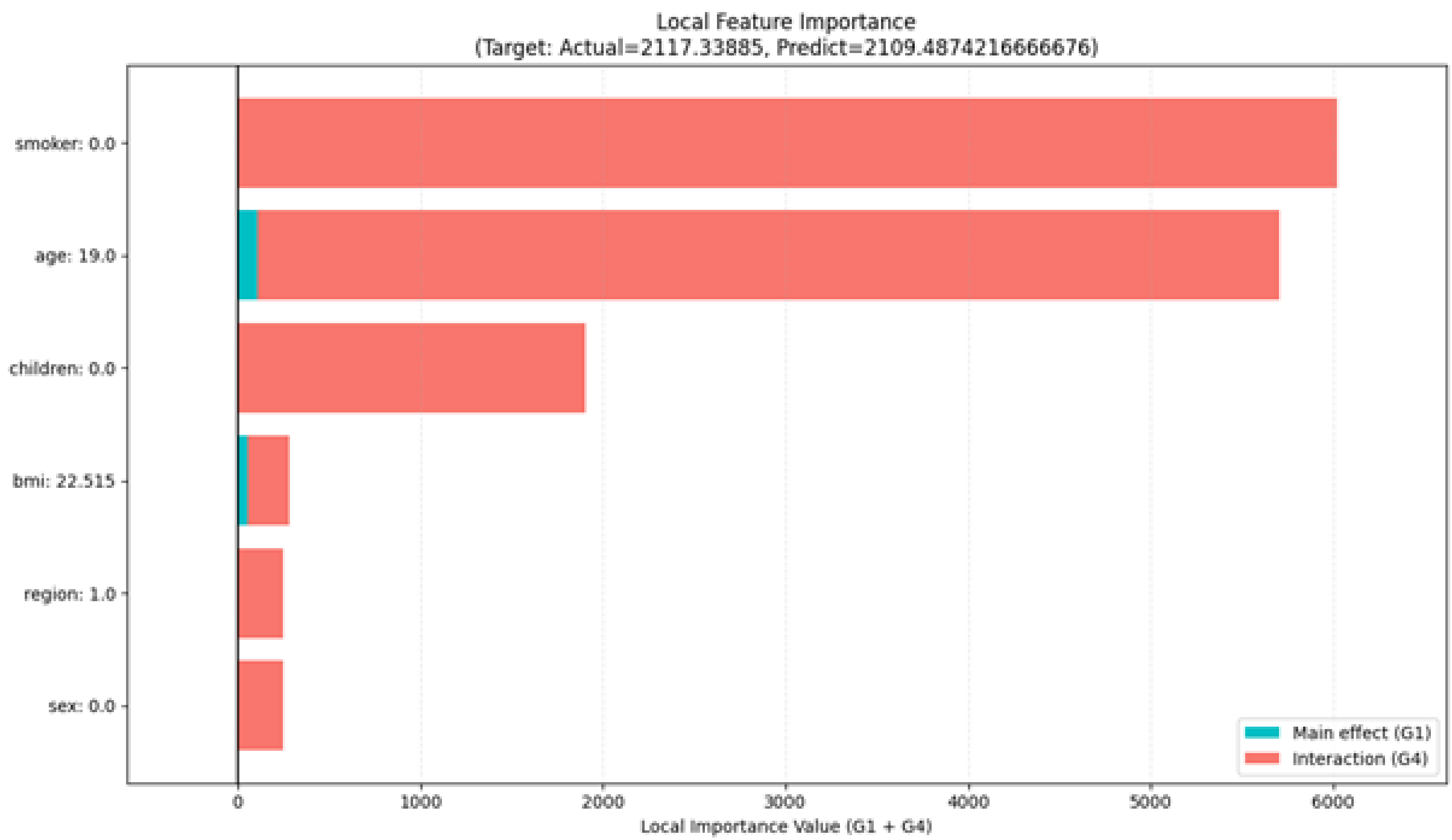

**Fig. 3.** Local feature importance for medical insurance dataset (*G1* vs. *G4* decomposition).

## Identification of regulatory mechanisms

Figure 4 shows the results of the regulatory-effect analyses**.** A unique advantage of localized CBFI is its ability to identify "regulatory mechanisms," where certain features suppress or attenuate the influence of others.

- **Suppressor Identification:** In the wine dataset, total sulfur dioxide showed a significant *G4* value (0.21), but its actual impact on the prediction probability (drop) was negative (-0.02).
- **Validation through Partial Dependence:** Analysis of the probability curves revealed that the high-quality classification probability consistently increased when total sulfur dioxide was removed (neutralized). Unlike additive models that treat all contributions as purely additive, CBFI explicitly identifies this variable as a "regulator" that suppresses the predicted probability, providing a deeper understanding of the model's nonlinear logic.

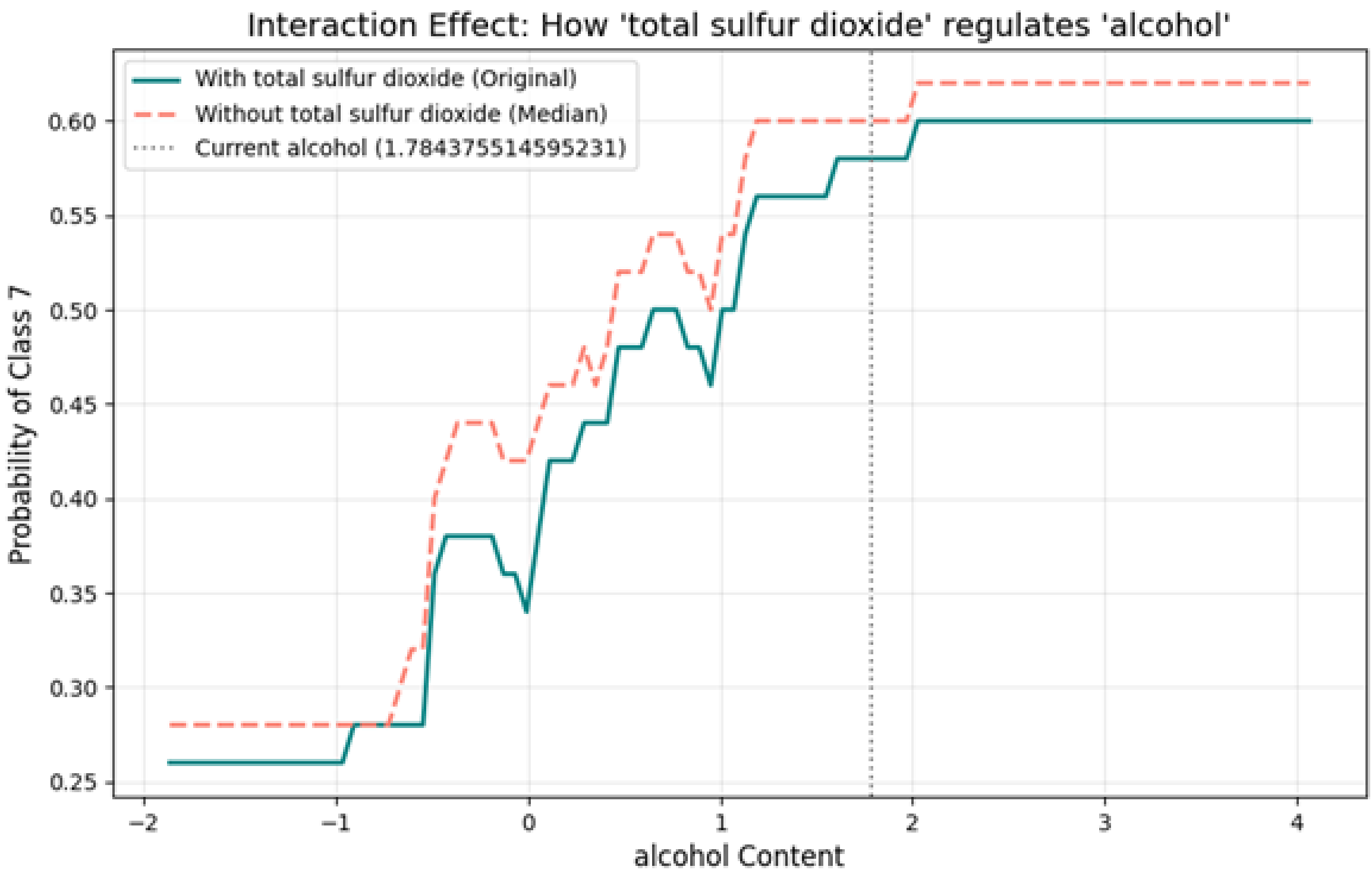

**Fig. 4.** Regulatory effect analysis: How total sulfur dioxide regulates alcohol.

## Explanatory fidelity and structural restoration

We conducted a fidelity test to evaluate how accurately the framework represented the decision structure of the model by sequentially removing features based on their importance rank. Figures 5 and 6 present the results of the fidelity test on the wine and insurance data.

- **Robustness of Feature Combinations (Wine Data):** Similarly, in the wine quality sample (#18), although SHAP and LIME showed steeper initial drops, the drop curve of CBFI deepened as more features were removed. At $k = 5$, CBFI reached the lowest correct class probability (0.06) among all the methods. This proves that localized CBFI is more effective at identifying the "essential set of features" required to sustain the model's decision rather than just identifying features with high marginal probability shifts.
- **Restoration of Interdependent Logic (Insurance Data):** In Sample #6 of the insurance dataset, SHAP showed an immediate shift in prediction upon removing the top-ranked feature ($k = 1$). However, localized CBFI maintained a stable prediction at $k = 1$ and exhibited a sharp drop only at $k = 2$, where the corresponding interaction pair was completely removed. This demonstrates that CBFI correctly identifies the "binding requirement" of synergistic features, where the decision structure only breaks down when the entire interaction is neutralized.

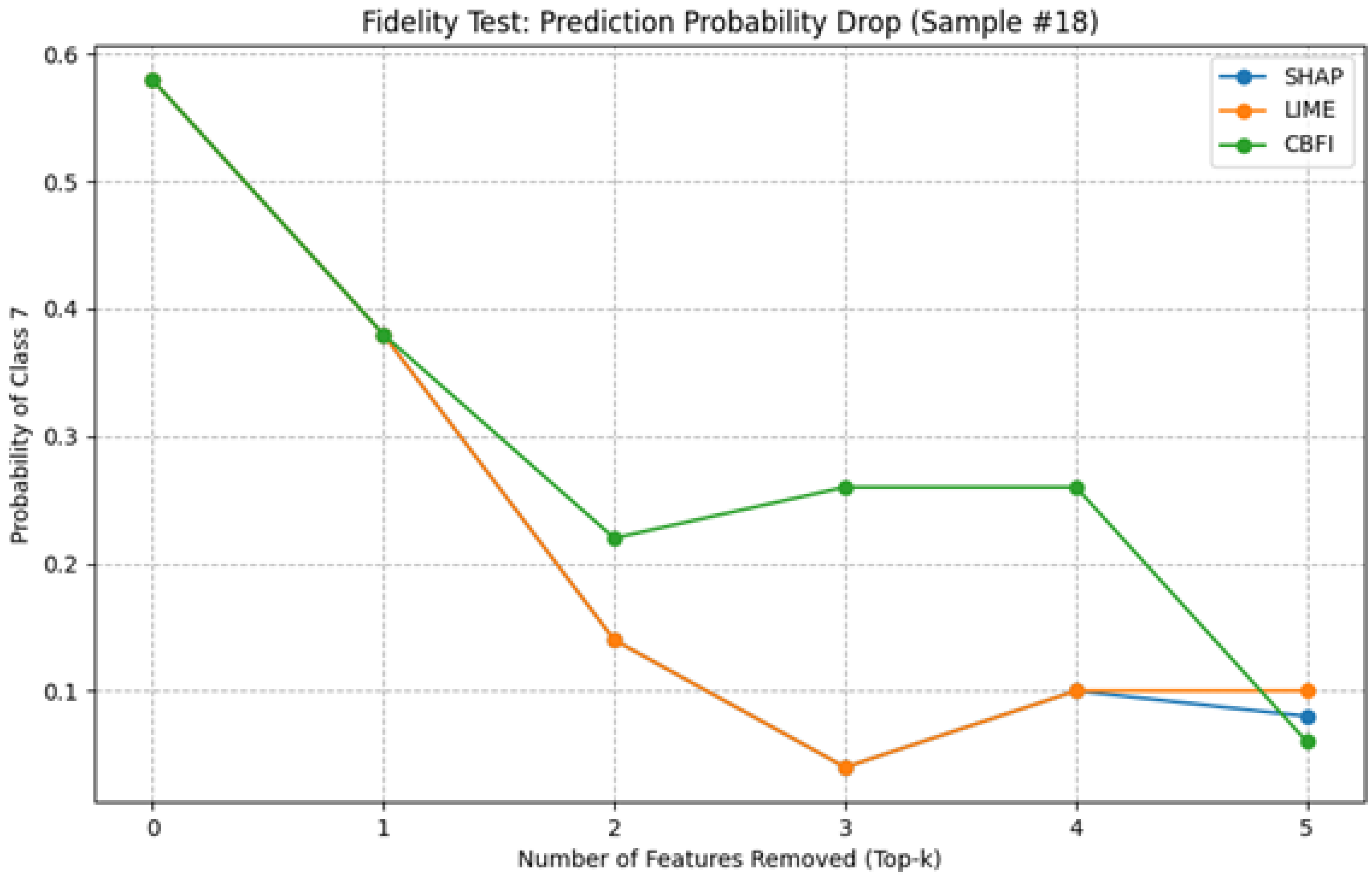

**Fig. 5.** Fidelity test: Prediction probability drop (Wine quality sample #18).

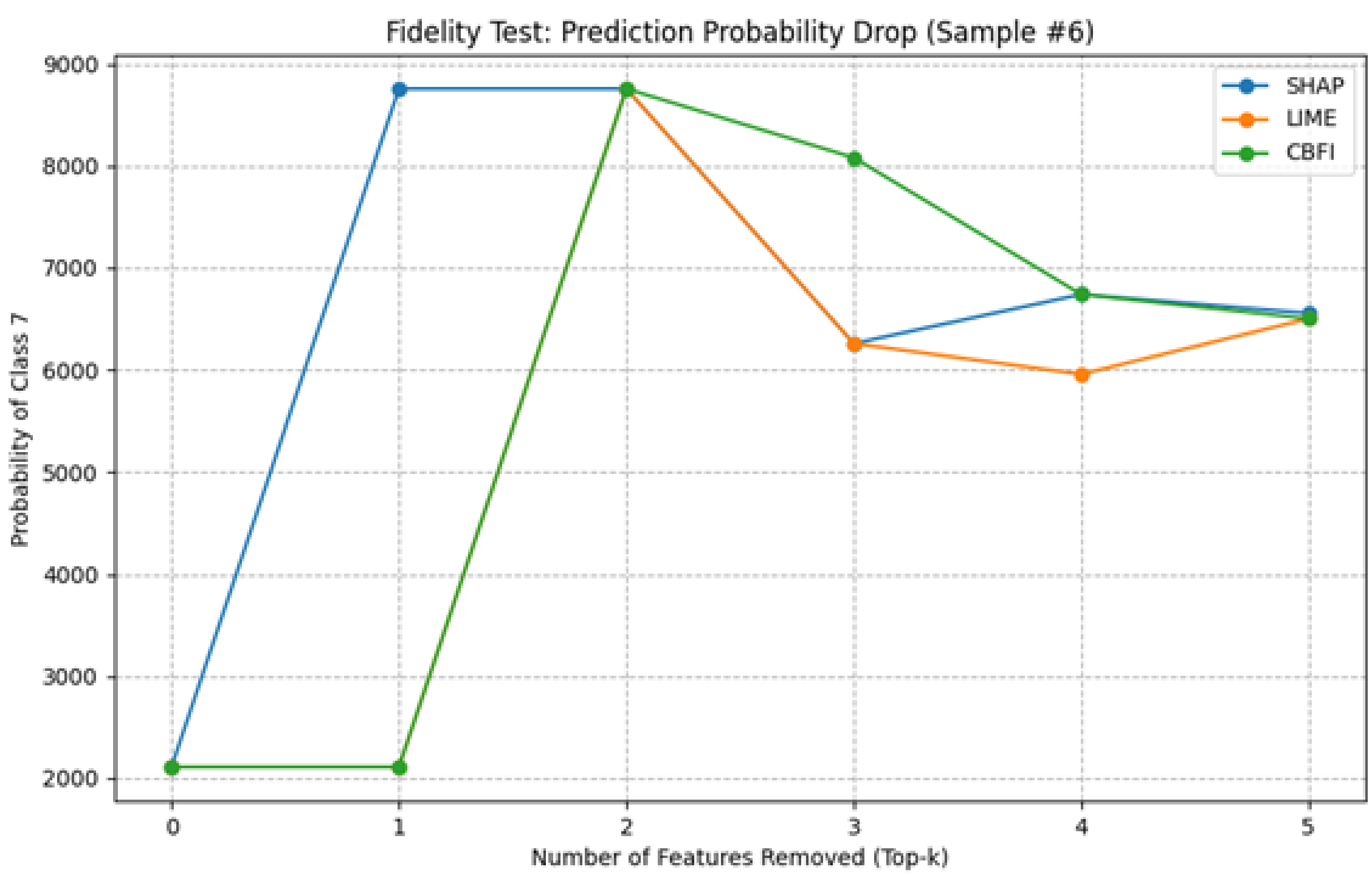

**Fig. 6.** Fidelity test: Prediction value shift (Medical insurance sample #6).

To further investigate how these interaction dependencies organize the structural dependency patterns of the model, we constructed a feature interaction graph based on the aggregated pairwise interaction strengths identified by CBFI (Figure 7). The resulting network reveals a clear structural hierarchy among the input variables. In particular, age and smoker emerged as the dominant interaction hubs mediating a large portion of the interaction pathways within the model. Strong connections, such as age–smoker, age–children, and smoker–BMI indicate that predictive signals are largely produced through synergistic feature combinations rather than independent feature contributions. These interactions suggest that the model captures realistic dependency patterns in which health-related risk factors reinforce each other to determine the final prediction. In contrast, variables such as sex and region appeared as peripheral nodes with relatively weak interaction connectivity, implying that their influence was primarily additive rather than structurally interactive. This interaction topology provides a structural view of the decision-making process of the model and illustrates how predictive signals propagate through interconnected feature relationships. Overall, the graph demonstrates that CBFI not only identifies influential variables but also restores the interaction structure underlying the model's predictive logic.

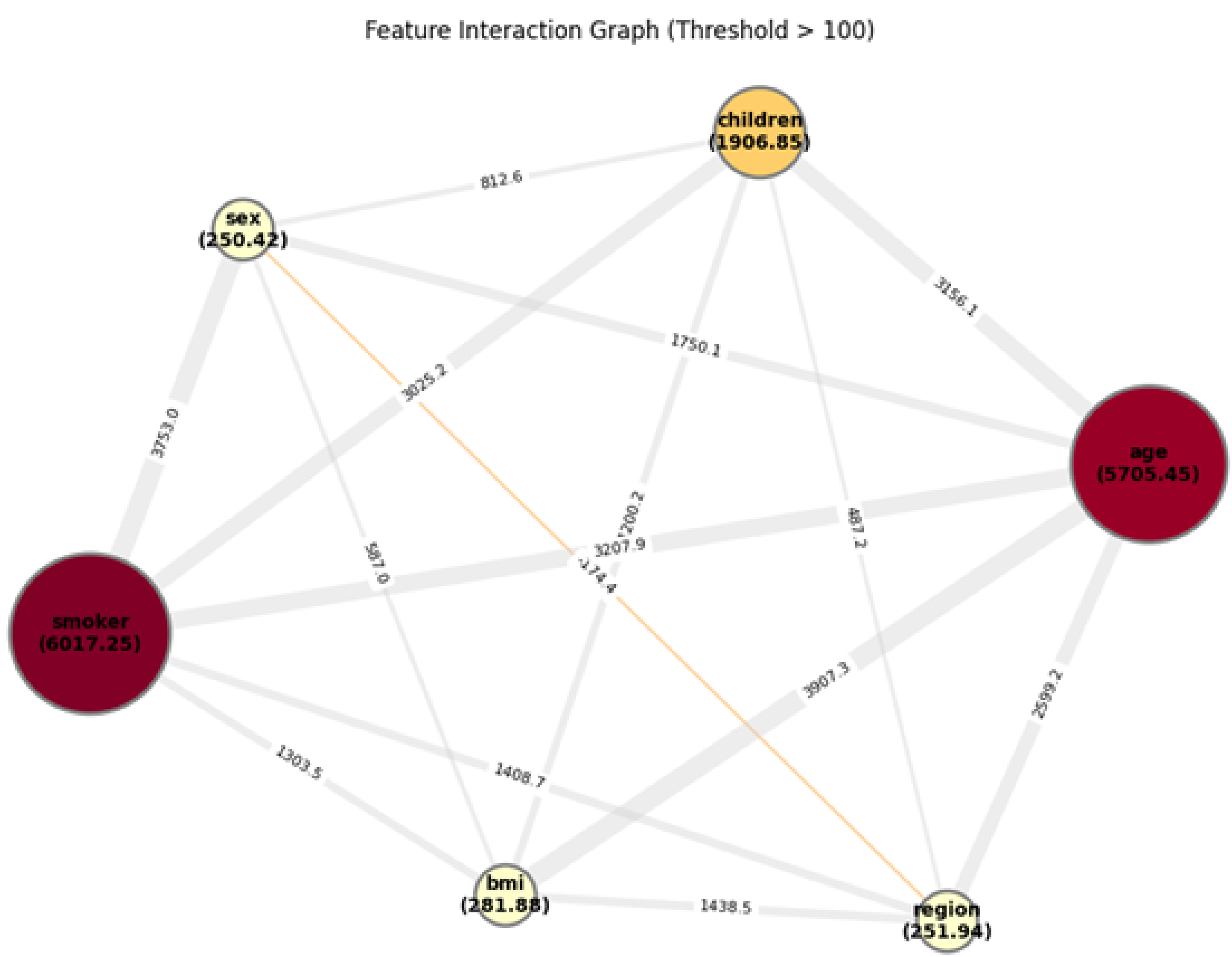

**Fig. 7.** Exposed latent interaction structure using CBFI (Insurance dataset). In the graph, nodes represent features, and larger and darker nodes indicate stronger main effects. Edges represent interactions between features, and thicker edges indicate stronger interactions. The orange lines denote negative interactions.

## Empirical validation of interaction structure identification: Tests on synthetic ground truth

To rigorously verify whether localized CBFI can accurately expose the structural dependency patterns learned by a model, we conducted a "Structural Ground Truth Identification" experiment using diverse synthetic scenarios. For this validation, we employed a RandomForestRegressor, an ensemble model capable of capturing complex nonlinear feature dependencies through hierarchical tree splits. Unlike traditional importance attribution, this experiment focuses on the disentanglement of main effects (*G1*) and interaction-driven contributions (*G4*) across five high-dimensional scenarios.

### Multi-faceted structural scenarios

We designed five distinct scenarios to evaluate the framework's robustness against various numerical mechanisms:

- **Additive ($y = \Sigma x_i$):** A baseline to ensure the framework avoids false interaction detection.
- **XOR ($y = XOR(x_1, x_2)$):** A logical interaction where individual features carry zero marginal information, testing the ability to recover **pure interaction necessity**.
- **Multiplicative ($y = 10 \times x_1 \times x_2 \times x_3$):** A test for identifying high-order synergistic interactions.
- **Suppressor ($y = 2\times x_1 - 5 \times x_1 \times x_2$):** A scenario where $x_2$ regulates the impact of $x_1$, requiring the detection of **negative interaction ($G4 < 0$)**.
- **Conditional ($y =$ *if* $x_5 > 0$ then $x_1 x_2$ else $x_1$):** A test for localized, context-dependent structural activation. For samples where x5 > 0, *G4* activation increased sharply, whereas interaction signals disappeared in the additive regime (x5 ≤ 0)

### Summary of findings

As summarized in Table 3 and the detailed numerical results in Appendix A4, the localized CBFI framework successfully distinguished between independent effects and synergistic dependencies, aligning precisely with the predefined ground truths.

Notably, in the XOR scenario, additive attribution frameworks such as SHAP distributed importance across individual features, despite the absence of meaningful standalone effects. In stark contrast, localized CBFI successfully suppressed *G1* to near-zero (<0.01) and isolated the interaction-driven dependency through a dominant *G4* indicator, precisely recovering the underlying structural necessity.

Furthermore, Localized CBFI exhibited substantially stronger interaction activation signals than SHAP interaction values in multiplicative scenarios. These results validate that localized CBFI provides an objective and high-resolution **diagnostic map** of a model's internal decision mechanisms.

The discriminative power of the *G4* indicator across diverse numerical environments is visually summarized in Figure 8. As illustrated, localized CBFI maintains a near-zero baseline in the **Additive** scenario while exhibiting robust, high-magnitude signals in complex structures such as **XOR** and **Multiplicative** models. Particularly, the clear negative orientation in the **Suppressor** case further validates the framework’s ability to not only detect the presence of interactions but also to identify their specific functional roles, such as regulatory suppression.

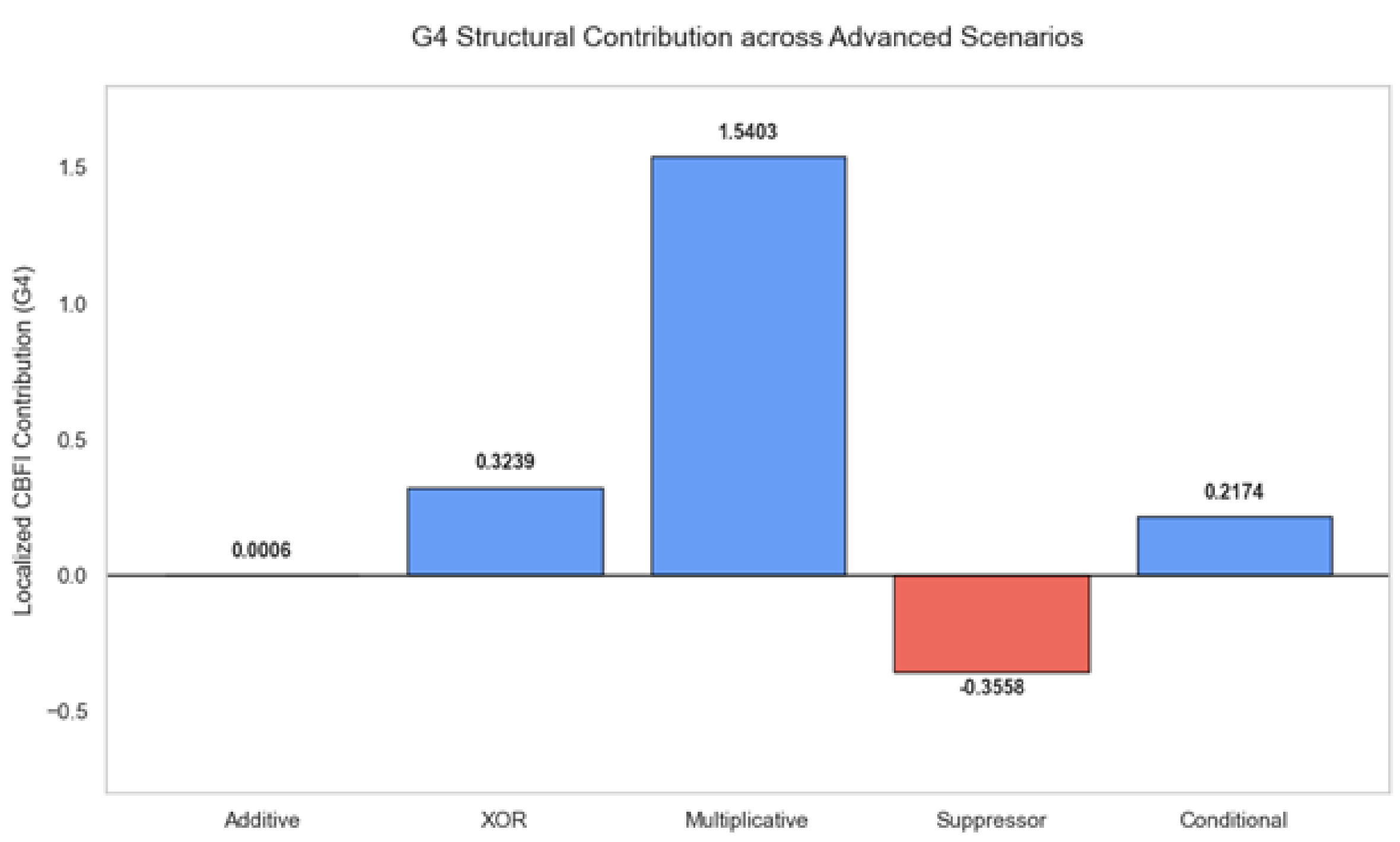

**Fig. 8.** Identification of Latent Interaction Structures.

To further demonstrate the superiority of localized CBFI, we compared its results with those of two widely used additive explanation models: SHAP and LIME. As shown in our synthetic experiments (**Table 4**), although SHAP and LIME identified $x_3$ and $x_4$ as significant, they provided only an **aggregated importance score** that conflates independent effects with interdependent synergies. Specifically, SHAP assigned its highest score to $x_3$ (0.6321) but failed to clarify whether this was derived from $x_1$ alone or from its coupling with $x_4$.

In contrast, localized CBFI explicitly **decoupled** these effects. By assigning a dominant *G4* score to the $x_3$-$x_4$ pair (0.5807 in basic interaction; **1.5403 in multiplicative cases**), our framework revealed that the influence of these features is fundamentally **structural and context-dependent**. This clear taxonomic separation allows practitioners to understand not just 'which' features are important but 'how' they function within the model's logic—a critical distinction that additive models inherently lack.

## Multi-domain validation

To evaluate the generalizability and robustness of the localized CBFI methodology, we conducted extensive experiments across six benchmark datasets spanning various domains, including finance, healthcare, and real estate, and six different machine learning architectures, including tree-based models and deep neural networks. Further details are provided in the appendix.

### Consistency across model architectures

Table 5 summarizes the mean values of the proposed metrics—main effect (*G1*), interaction (*G4*), and overall importance—along with the runtime for each model-dataset pair. These results demonstrate that localized CBFI consistently quantifies interactions regardless of the complexity of the underlying model. Notably, in complex regression tasks, such as Ames Housing and Bike Sharing, interaction (*G4*) often exhibited a higher magnitude than main effect (*G1*), revealing that model predictions in these domains are heavily driven by feature synergies rather than isolated effects. This indicates that the predictive mechanism of these models is primarily governed by interactions rather than by independent feature contributions.

### Comparative fidelity and alignment with SHAP

We further analyzed the alignment between the CBFI importance (*G1* + *G4*) and established XAI metrics. In the classification tasks (e.g., Adult and Diabetes datasets), our importance scores showed a strong correlation ($R > 0.6$) with the SHAP values, serving as a sanity check for the proposed method. Crucially, the observed divergence from SHAP in complex regression tasks is not a methodological discrepancy but rather empirical evidence of the structural distortion inherent in additive frameworks. The low correlation coefficients ($R$) highlight specific data regions where additive approximations mask critical interactions, which localized CBFI successfully rectifies by restoring latent interaction logic.

To further examine this divergence, we analyzed the relationship between the decomposed components of the proposed framework and the feature attributions produced by SHAP. Whereas the independent main effect component (*G1*) showed moderate agreement with the SHAP feature importance values, the interaction component (*G4*) often exhibited weaker or inconsistent correlations with the additive attributions. This behavior is expected because additive explanation models distribute predictive influence across individual features, implicitly absorbing nonlinear dependencies into feature-level importance scores. In contrast, the localized CBFI framework explicitly isolates the contribution arising from feature combinations, enabling the identification of latent interaction structures that govern how multiple variables jointly influence model predictions. Consequently, the divergence from SHAP reflects the ability of the proposed method to reveal structural dependency patterns that remain hidden in additive attribution models.

### Computational Feasibility for Real-time Systems

The average runtime per instance remained under 1.0 s for most datasets, even when explaining high-dimensional models such as TabNet or ensembles such as CatBoost. This efficiency highlights the practicality of localized CBFI for integration into database systems that require near-real-time explanation generation, without the prohibitive computational costs typically associated with high-order interaction discovery.

## Comparative analysis with interaction-specific metrics

To rigorously evaluate the distinctive advantages of localized CBFI, we conducted a comparative analysis with established interaction-specific metrics: SHAP Interaction Value and Friedman's H-statistic. This comparison focuses on how each metric identifies and quantifies non-linear dependencies.

### Experimental design

The comparison was performed using a synthetic dataset where the ground truth is explicitly defined as $y = x_1 + x_2 + 5(x_3 \times x_4) + \varepsilon$. This setup allows us to evaluate the metrics based on their ability to:

1. Isolate pure interactions: Distinguishing between main effects and synergistic couplings.
2. Numerical Robustness: Avoiding the "additive bias" where interaction effects are leaked into individual feature importance.
3. Diagnostic Granularity: Providing directional information (synergy vs. suppression) beyond simple magnitude.

### Experimental Results

The quantitative comparison reveals distinct differences in how each metric discriminates between additive effects and synergistic interactions. To ensure a fair comparison across different mathematical scales, we introduce the **Sensitivity** metric, defined as the ratio of the signal (interactive features) to the noise (additive features).

As demonstrated in Table 6, while all metrics correctly identify the presence of interaction, their **discriminative power** varies significantly. SHAP Interaction values remain relatively suppressed (0.0503) with a sensitivity of 41.9x, suggesting that a substantial portion of the interaction effect is still absorbed into individual feature attributions. Friedman's H-statistic confirms the structural interaction at a global level ($H^2$ = 0.1445), but exhibits the lowest sensitivity (13.9x) due to the inherent variance in partial dependence calculations.

In contrast, **localized CBFI (*G4*)** achieves an overwhelming sensitivity of **1,952.7x**. It not only captures the high-magnitude signal (1.1716) corresponding to the injected synergistic coefficient but also maintains the lowest noise level (0.0006) for purely additive features.

### Evaluation and interpretation

The exceptional sensitivity of localized CBFI provides several critical advantages for model diagnostics:

- **Superior Signal-to-Noise Ratio (SNR):** The 1,952.7x sensitivity ratio effectively eliminates the ambiguity often found in traditional metrics. This ensures that the **exposed structural dependency patterns** are not artifacts of numerical noise but represent the true functional coupling learned by the model.
- **Resolution of Additive Bias:** The divergence in sensitivity highlights a fundamental flaw in traditional allocation-based models (e.g., SHAP), which tend to "smooth over" complex interactions. By **disentangling interaction-driven contributions**, localized CBFI prevents the dilution of synergistic effects into main effect scores, providing a more precise map of the model's internal mechanisms.
- **Mathematical Objectivity and Robustness:** The extreme contrast between the signal and noise levels demonstrates that the *G4* indicator is highly specialized for **identifying latent interaction structures**. This high-resolution identification allows researchers to distinguish between trivial correlations and critical synergistic partnerships that govern the model's final predictions.

## Numerical stability and convergence analysis

In this section, we analyze the numerical stability and convergence characteristics of the localized interaction metric (*G4*) as a function of the number of samples ($n_{samples}$). Given that sampling-based XAI methodologies inherently involve stochastic processes, ensuring the reproducibility and reliability of the explanations is of paramount importance. To this end, we conducted a sensitivity analysis across six benchmark datasets and various machine learning algorithms.

### Convergence of localized interactions

Figure 9 illustrates the convergence behavior of the mean *G4* value as $n_{samples}$ increases from 10 to 500. The experimental results demonstrate a consistent pattern across all datasets and models, where the fluctuations in the *G4* value diminish and reach a plateau as the sample size increases.

- Identification of the Elbow Point: Most models reached a stable convergence state within the range of $n$ = 100–200. Tree-based models, such as XGBoost and random forest, exhibit particularly rapid convergence, achieving stable importance scores even with relatively small sample sizes.
- Scalability to High-Dimensional Data: Even in the Ames Housing dataset, which features a high-dimensional feature space, the *G4* values formed stable convergence curves without divergence. This confirms that the proposed localized CBFI methodology operates robustly within complex nonlinear decision spaces.

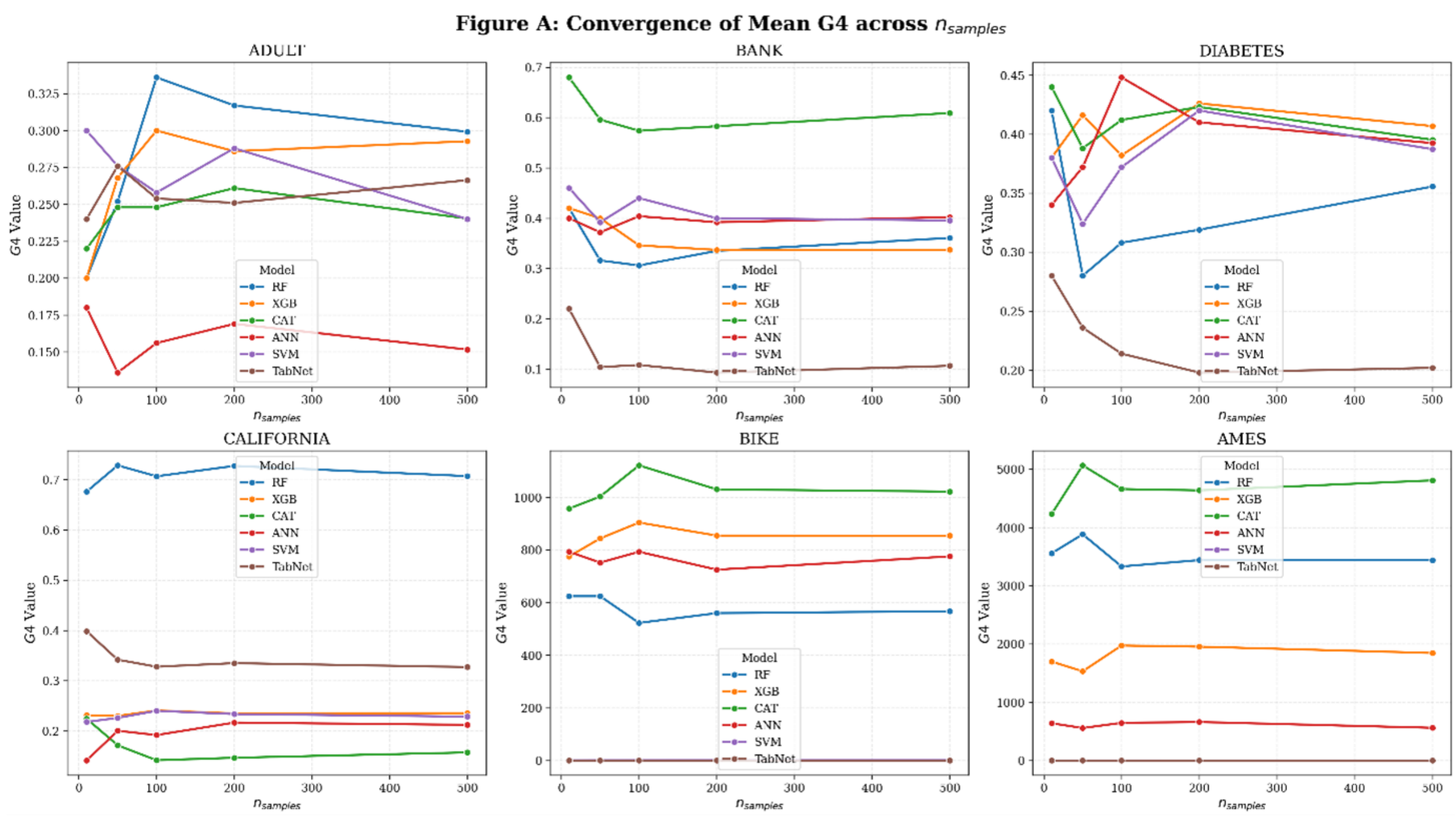

**Fig. 9.** Mean *G4* convergence curves.

### Statistical stability via coefficient of variation (CV)

Figure 10 presents the coefficient of variation (CV) of the *G4* metric across repeated trials, quantifying the impact of sampling noise on the final explanation.

- Analysis of Error Reduction: Consistent with the theoretical expectations for the Monte Carlo estimation, the *CV* decreased at a rate of approximately $1/\sqrt{n}$. Although the variation was high at $n$ = 10 (averaging between 0.4 and 0.6), it stabilized significantly as it approached the $n$ = 100 mark, often falling below the 10% threshold (indicated by the red dotted line).

- Algorithm-Specific Robustness: Models with smoother decision boundaries, such as SVM and MLP, exhibit exceptionally low CV values. Conversely, more complex models such as TabNet require a larger number of samples ($n \geq 200$) to maintain a stability level of approximately 5%, reflecting the intrinsic complexity of their internal logic.

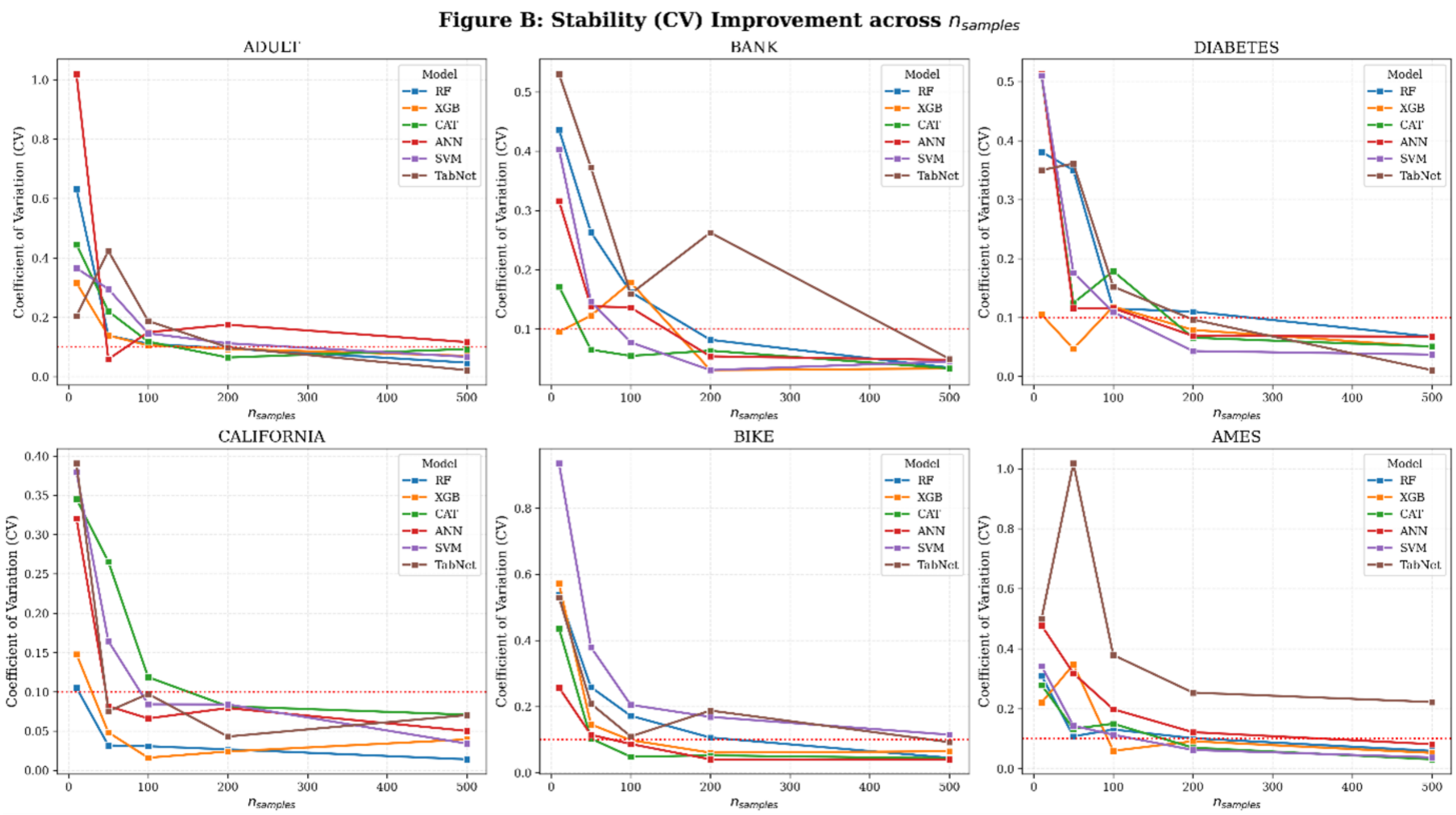

**Fig. 10.** Stability (CV) improvement plots.

### Justification of the experimental setting (*n* = 100)

The empirical data indicate that the difference in the mean *G4* value between $n$ = 100 and $n$ = 500 was less than 5% in most cases. This suggests that $n$ = 100 serves as an optimal "sweet spot" that ensures statistical significance while maintaining computational efficiency in large-scale database environments. Consequently, the primary experiments and comparative analyses were conducted using $n$ = 100 as the validated standard.

## Discussion

The localized CBFI framework proposed in this study aims to overcome the structural limitations of the existing additive explanation models by explicitly decomposing feature interactions at the individual instance level. In this section, we discuss the academic and practical implications of our findings from six key perspectives based on the experimental results.

### Breaking the additive trap: From aggregated importance to structural decomposition

Standard XAI models, such as SHAP and LIME, typically present feature contributions in an additive form. While this provides computational convenience, it carries the risk of misrepresenting the "source" of contributions in complex datasets where feature interactions are dominant (Alvarez-Melis & Jaakkola, 2018). The results from the medical insurance dataset clearly illustrate this limitation. When SHAP assigns a high importance score to "smoker," it fails to distinguish whether this stems from the independent effect of smoking or its synergistic amplification with "age" or "BMI." In contrast, localized CBFI explicitly separates these into *G1* and *G4* components, effectively restoring the model's decision structure as a "logical structure" rather than a mere "sum of values" (Oh, 2022). This signifies a paradigm shift in explaining black-box models—moving beyond "how much" a feature contributes to revealing "in what manner" it contributes.

## Decoding inhibitory interactions: Significance of regulatory mechanisms

The majority of existing research on interactions has focused on "positive synergy," where the combined effect of two variables exceeds the sum of their individual parts. However, the case of "total sulfur dioxide" in our wine quality experiment highlights the critical importance of "inhibitory or regulatory mechanisms" (Cohen et al., 2013). Despite recording a high interaction (*G4*) value, this variable was found to act as a regulator that suppresses the probability of a high-quality classification. The ability to identify such "negative interactions" is of paramount value in precision domains such as fine chemistry, drug formulation, and manufacturing optimization. By quantitatively identifying the components that interfere with the effects of others, the framework provides prescriptive guidance for performance improvement, rather than just a passive description of the status quo.

## Providing a stable anchor in explanatory conflict zones

A persistent challenge in the XAI community is the inconsistency of explanations, as SHAP and LIME often provide conflicting results for the same data and model. This "explanatory divergence" was observed in our insurance dataset experiment, where the correlation between SHAP and LIME importance scores dropped to -0.6029. Amidst this confusion, localized CBFI maintained a robust and consistent stance, showing high correlations with both SHAP (in absolute terms, 0.93) and LIME (0.75). This demonstrates that our framework is not swayed by specific algorithmic assumptions of local linear surrogates or game-theoretic distributions but instead captures the structural truth of the underlying model (Kohavi, 1996). Such stability is vital for establishing trust in sectors such as finance and healthcare, where explanatory reliability is not negotiable.

## Enhancing practical utility through absolute interpretation

Normalized probability values or abstract importance scores are often counterintuitive for domain experts when making real-world decisions. A distinctive advantage of localized CBFI is its output in "absolute units" such as actual insurance charges ($) or direct probability shifts (%). For instance, an insurance actuary's ability to explain to a client that "quitting smoking will specifically save $6,000 by neutralizing the synergy with age" is far more persuasive than any relative score (Moro, Cortez & Rita, 2014). By preserving the original scale of the output while isolating interactions, localized CBFI serves as a key driver for enhancing the social acceptance of AI models and implementing responsible AI in high-stakes decision-making environments.

## Practical implications and methodological robustness

The results of the stability and convergence analyses provide several critical academic insights into how localized CBFI balances explainability with statistical reliability.

### Consistency with theoretical framework

The observed $1/\sqrt{n}$ decay pattern of CV confirms that our methodology strictly adheres to the fundamental principles of Monte Carlo estimation (Smith et al., 1988). This signifies that localized CBFI is grounded in rigorous statistical theory rather than mere heuristics. Crucially, the decision structure defined in Figure 1, which distinguishes between *G1* (main effect) and *G4* (interaction), demonstrates remarkable resilience to sampling noise. The consistent signs and magnitudes of these metrics, even under stochastic sampling, validate the robustness of the error-reduction-based attribution mechanism.

### Efficiency–precision trade-off: The *n* = 100 "Sweet Spot"

The adoption of $n = 100$ in our primary experiments was a strategic choice based on the empirical evidence. Our convergence curves indicate that quintupling the sample size from $n = 100$ to $n = 500$ yields a marginal information gain of less than 5% while increasing the computational overhead linearly. In the context of large-scale database systems, in which providing near-real-time explanations is a primary requirement, this "sweet spot" ensures that the methodology remains practically viable without sacrificing the integrity of the interaction discovery process.

### Impact of model architecture on explanation stability

An intriguing finding is the relationship between model nonlinearity and convergence speed. Models with smoother decision boundaries, such as SVMs, exhibited immediate stability with minimal samples. In contrast, models with "rugged" or complex boundaries, such as TabNet and XGBoost, required higher sampling density to achieve comparable stability. This suggests that localized CBFI effectively modulates its explanatory resolution to match the intrinsic complexity of the underlying model, further reinforcing its validity as a model-agnostic tool for auditing black-box systems.

## Interpreting methodological versatility and interaction discovery

The multidomain validation results provide several profound insights into the underlying mechanics of localized CBFI and its departure from traditional additive explanation frameworks.

### Beyond the additive fallacy: significance of *G4* dominance

A critical observation in our experiments is the relative magnitude of interaction (*G4*) compared to main effect (*G1*), particularly in complex regression tasks, such as Ames Housing. Traditional XAI methods such as SHAP operate on an additive assumption, where feature effects are treated as independent contributions to a sum. However, our findings suggest that in high-stakes predictive modeling, the "whole" is often significantly different from the sum of its parts. The dominance of *G4* in these domains indicates that localized CBFI successfully uncovers nonlinear synergies that additive models implicitly smooth over. This ability to explicitly separate interaction from isolated effects prevents the "dilution" of feature importance, providing a more granular map of the model's decision-making process.

### Decoding divergence: Why low correlation with SHAP is a success

While the high correlation with SHAP in the classification tasks serves as a baseline validation (sanity check), the observed divergence ($R < 0.3$) in the complex regression domains is perhaps the most significant finding. This lack of correlation should not be interpreted as a methodological inconsistency; rather, it highlights a fundamental shift in explanatory perspective. SHAP seeks to distribute the "payout" of a prediction among features, whereas localized CBFI seeks to identify which feature combinations actively reduce predictive error. In environments in which feature interactions are the primary drivers of variance, localized CBFI captures distinct signals that additive approximations are mathematically structured to miss. Localized CBFI is a necessary complementary tool for auditing models where nonlinearity is a feature and not a nuisance.

### Diagnostic utility of negative effect (*G1*)

The emergence of negative *G1* values in certain regression instances offers unique diagnostic capability. Whereas conventional importance metrics are typically non-negative or represent the direction of change, a negative *G1* in the CBFI framework specifically identifies a feature that, in its current local context, increases the model's error. This allows data scientists to identify not just "important" features but "harmful" ones that might be misleading the model in specific regions of the data space. Such insights are invaluable for model debugging and ensuring the fairness and reliability of automated systems in sensitive domains, such as healthcare and finance.

### Scalability as a catalyst for explainable database systems (XDB)

Finally, the consistent sub-second runtime across all 36 model-dataset combinations reinforces the practical viability of our approach. Several interaction-aware XAI methods suffer from exponential complexity as the number of features increases. By utilizing a contribution-based sampling strategy, localized CBFI maintains linear scalability with respect to the number of features. This efficiency is a critical enabler for the next generation of "Explainable Databases," where the system must be capable of generating on-the-fly explanations for individual query results without degrading transaction performance (De Cock, 2011).

Beyond numerical robustness, our framework shifts the paradigm from "approximated additive summaries" to "precise structural diagnostics." The computational efficiency of localized CBFI enables the real-time restoration of decision structure within large-scale database systems. This allows practitioners to audit complex black-box models without the structural fidelity loss inherent in traditional additive approximations (e.g., SHAP or LIME). By explicitly decomposing error reduction into main effects and interactions, the proposed framework provides a rigorous foundation for rectifying the structural distortions that have long compromised the reliability of local explanations.

## Conclusions

This study introduced the localized CBFI framework to decompose model predictions into independent main effect (*G1*) and explicit interaction (*G4*) at the individual instance level. Unlike additive explanation methods, such as SHAP and LIME, the proposed approach isolates the structural contributions arising from feature combinations. The experimental results demonstrate that localized CBFI not only maintains a strong alignment with established attribution methods but also reveals interaction-driven mechanisms, such as suppression effects and synergistic dependencies, which are often obscured in additive explanations. By reconstructing instance-specific interaction logic and expressing contributions on interpretable absolute scales, such as probability or monetary changes, the framework provides a more faithful and practically meaningful interpretation of complex machine learning models. More importantly, the framework reframes explainability from importance attribution to exposing structural dependency patterns, offering a new perspective on how machine learning models should be interpreted. The library and experimental code for Localized CBFI are available at https://github.com/dkumango/Local_CBFI/.

Despite the demonstrated advantages, this study has a few limitations that should be acknowledged. First, as a perturbation-based method relying on Monte Carlo sampling, the localized metrics inherently possess a degree of stochastic variance. Although empirical stability was confirmed at $n=100$, exact deterministic attributions cannot be guaranteed. Second, while the framework effectively identifies pairwise (two-way) synergistic and suppressive effects, quantifying higher-order interactions (e.g., three-way or beyond) remains computationally challenging due to the combinatorial nature of complex feature subsets. Finally, the current scope of empirical evaluation is restricted to tabular data environments, leaving its efficacy on unstructured data unverified.

Future studies should focus on improving the computational efficiency of high-dimensional tabular data and extending the framework to a broader range of learning architectures. In particular, evaluating its compatibility with advanced tabular deep learning models and gradient boosting systems, such as XGBoost and CatBoost, will further validate its general applicability. In addition, we plan to develop human-centered visualization tools to explore interactions and investigate the applicability of localized contribution decomposition in multimodal domains, including text and image analyses.

## Acknowledgements

N/A

## References

Adadi A, Berrada M. 2018. Peeking inside the black-box: a survey on explainable artificial intelligence (XAI). *IEEE Access* 6:52138–52160.

Alvarez-Melis D, Jaakkola TS. 2018. On the robustness of interpretability methods. In: *Proceedings of the 32nd International Conference on Neural Information Processing Systems (NIPS)*. pp. 4511–4520.

Arık SÖ, Pfister T. 2021. TabNet: Attentive interpretable tabular learning. In: *Proceedings of the 35th AAAI Conference on Artificial Intelligence*. pp. 6679–6687.

Arrieta AB, Díaz-Rodríguez N, Del Ser J, Bennetot A, Tabik S, Barbado A, García S, Gil-López D, Molina R, Benjamins R, Chatila F, Herrera F. 2020. Explainable Artificial Intelligence (XAI): Concepts, taxonomies, opportunities and challenges toward responsible AI. *Information Fusion* 58:82–115.

Breiman L. 2001. Random forests. *Machine Learning* 45:5–32.

Chen T, Guestrin C. 2016. XGBoost: A scalable tree boosting system. In: *Proceedings of the 22nd ACM SIGKDD International Conference on Knowledge Discovery and Data Mining (KDD)*. pp. 785–794.

Cohen J, Cohen P, West SG, Aiken LS. 2013. *Applied multiple regression/correlation analysis for the behavioral sciences*. New York: Routledge.

Cortes C, Vapnik V. 1995. Support-vector networks. *Machine Learning* 20:273–297.

Cortez P, Cerdeira A, Almeida F, Matos T, Reis J. 2009. Modeling wine preferences by data mining from physicochemical properties. *Decision Support Systems* 47:547–553.

De Cock D. 2011. Ames, Iowa: Alternative to the Boston housing data as an end of semester regression project. *Journal of Statistics Education* 19(3):1–15.

Friedman JH, Popescu BE. 2008. Predictive learning via rule ensembles. *The Annals of Applied Statistics* 2:916–954.

Gilpin LH, Bau D, Yuan BZ, et al. 2018. Explaining explanations: An overview of interpretability of machine learning. In: *Proceedings of the IEEE 5th International Conference on Data Science and Advanced Analytics (DSAA)*. pp. 80–89.

Grinsztajn L, Oyallon E, Varoquaux G. 2022. Why do tree-based models still outperform deep learning on typical tabular data? In: *Proceedings of the 36th International Conference on Neural Information Processing Systems (NeurIPS)*. pp. 507–520.

Guidotti R, Monreale A, Suggieri S, Turini F, Giannotti F, Pedreschi D. 2018. A survey of methods for explaining black box models. *ACM Computing Surveys* 51:1–42.

Kohavi R. 1996. Scaling up the accuracy of naive-bayes classifiers: a decision-tree hybrid. In: *Proceedings of the 2nd International Conference on Knowledge Discovery and Data Mining (KDD)*. pp. 202–207.

Lantz B. 2013. *Machine Learning with R*. Birmingham: Packt Publishing Ltd.

Lundberg SM, Erion G, Chen H, et al. 2020. From local explanations to global understanding with explainable AI for trees. *Nature Machine Intelligence* 2:56–67.

Lundberg SM, Lee SI. 2017. A unified approach to interpreting model predictions. In: *Proceedings of the 31st International Conference on Neural Information Processing Systems (NIPS)*. pp. 4765–4774.

Miller T. 2019. Explanation in artificial intelligence: Insights from the social sciences. *Artificial Intelligence* 267:1–38.

Molnar C, Casalicchio G, Barch B. 2020. Quantifying model complexity via functional decomposition for better post-hoc interpretability. *Communications in Computer and Information Science*. Berlin: Springer. pp. 193–204.

Moro S, Cortez P, Rita P. 2014. A data-driven approach to predict the success of bank telemarketing. *Decision Support Systems* 62:22–31.

Oh S. 2022. Predictive case-based feature importance and interaction. *Information Sciences* 593:155–176.

Owen AB. 2013. *Monte Carlo theory, methods and examples*. Stanford: Stanford University.

Pace RK, Barry R. 1997. Sparse spatial autoregressions. *Statistics & Probability Letters* 33:291–297.

Prokhorenkova L, Gusev G, Vorobev A, Dorogush AV, Gulin A. 2018. CatBoost: unbiased boosting with categorical features. In: *Proceedings of the 32nd International Conference on Neural Information Processing Systems (NIPS)*. pp. 6638–6648.

Ribeiro MT, Singh S, Guestrin C. 2016. "Why should I trust you?": Explaining the predictions of any classifier. In: *Proceedings of the 22nd ACM SIGKDD International Conference on Knowledge Discovery and Data Mining (KDD)*. pp. 1135–1144.

Rudin C. 2019. Stop explaining black box machine learning models for high stakes decisions and use interpretable models instead. *Nature Machine Intelligence* 1:206–215.

Rumelhart DE, Hinton GE, Williams RJ. 1986. Learning representations by back-propagating errors. *Nature* 323:533–536.

Slack D, Hilgard S, Jia E, et al. 2020. Fooling LIME and SHAP: Adversarial attacks on post hoc explanation methods. In: *Proceedings of the AAAI/ACM Conference on AI, Ethics, and Society (AIES)*. pp. 180–186.

Smith JW, Everhart JE, Dickson WC, Knowler WC, Johannes RS. 1988. Using the ADAP learning algorithm to forecast the onset of diabetes mellitus. In: *Proceedings of the Annual Symposium on Computer Applications in Medical Care*. pp. 261–265.

Tsang MH, Liu D, Purushotham S, et al. 2018. Neural interaction transparency (NIT): Explaining neural networks via structuring interactions. In: *Proceedings of the 32nd International Conference on Neural Information Processing Systems (NIPS)*. pp. 5827–5837.

Wang D, Yang Q, Abdul A, Lim BY. 2019. Designing theory-driven user-centric explainable AI. In: *Proceedings of the CHI Conference on Human Factors in Computing Systems*. pp. 1–15.